\documentclass{article}
\pdfoutput=1
\usepackage{graphicx}
\usepackage{float}
\usepackage[final]{nips_2017}
\usepackage{hyperref}       % hyperlinks
\usepackage{url}            % simple URL typesetting
\usepackage{booktabs}       % professional-quality tables
\usepackage{nicefrac}       % compact symbols for 1/2, etc.
\usepackage{microtype}
\usepackage{caption} 
\usepackage{subcaption}
\usepackage{amsmath}
\DeclareMathOperator*{\argmax}{argmax}
\usepackage{algpseudocode}
\usepackage{algorithm}
\algnewcommand\algorithmicforeach{\textbf{for each:}}
\algnewcommand\ForEach{\item[ \algorithmicforeach]}     % microtypography
\makeatletter
\g@addto@macro\@floatboxreset{\centering}
\makeatother 
\title{Explaining hyperspectral imaging based plant disease identification: 3D CNN and saliency maps}

\author{
  Koushik Nagasubramanian\\
  Department of Electrical Engineering\\
  Iowa State University\\
  \texttt{koushikn@iastate.edu} \\
  \And Sarah Jones\\
  Department of Agronomy. \\
  Iowa State University\\
 \texttt{sejones2@iastate.edu} \\ 
 \And Asheesh K. Singh\\
  Department of Agronomy. \\
  Iowa State University\\
 \texttt{singhak@iastate.edu} \\
 \And
  Arti Singh\\
  Department of Agronomy. \\
  Iowa State University\\
 \texttt{arti@iastate.edu} \\
 \And
  Baskar Ganapathysubramanian\\
  Department of Mechanical Engineering \\
  Iowa State University\\
 \texttt{baskarg@iastate.edu} \\
 \And
  Soumik Sarkar\\
  Department of Mechanical Engineering \\
  Iowa State University\\
 \texttt{soumiks@iastate.edu} \\
} 
\begin{document}
\maketitle
\begin{abstract}
Our overarching goal is to develop an accurate and explainable model for plant disease identification using hyperspectral data. Charcoal rot is a soil borne fungal disease that affects the yield of soybean crops worldwide. Hyperspectral images were captured at 240 different wavelengths in the range of 383 - 1032 nm. We developed a 3D Convolutional Neural Network model for soybean charcoal rot disease identification. Our model has  classification accuracy of 95.73$\%$ and an infected class F1 score of 0.87. We infer the trained model using saliency map and visualize the most sensitive pixel locations that enable classification. The sensitivity of individual wavelengths for classification was also determined using the saliency map visualization. We identify the most sensitive wavelength as 733 nm using the saliency map visualization. Since the most sensitive wavelength is in the Near Infrared Region (700 - 1000 nm) of the electromagnetic spectrum, which is also the  commonly used spectrum region for determining the  vegetation health of the plant, we were more confident in the predictions using our model.  
\end{abstract}
\section{Introduction}
Deep Convolutional Neural Networks(CNN) have led to rapid developments in diverse applications such as object recognition, speech recognition, document reading and sentiment analysis within the industry and social media \citep{krizhevsky2012imagenet,waibel1989phoneme,lecun1998gradient,dos2014deep}. Recently, 3D-CNN models  have been used in classification of hyperspectral images for different objects of interest \citep{fotiadou2017deep,Chen2016,Li2017}. But now that we are trying to leverage CNN's for science applications, where simply prediction results will not be sufficient to trust  the decision making of our model \citep{lipton2016mythos}. Therefore, we need interpretability methods for visualizing the information learnt by the model. There are many emerging techniques for interpreting CNN \cite{montavon2017methods} and one of them is saliency map based visualization developed by {\citep{simonyan2013deep}. Here we show, an important science application of it in the domain of plant pathology.
\par 
Plant diseases negatively impact yield potential of crops worldwide, including soybean [Glycine max (L.) Merr.], reducing the average annual soybean yield by an estimated 11\% in the United States \citep{hartman2015compendium}. %From 2010 to 2014, soybean economic damage due to disease could have accounted for over an estimated \$23 billion US dollars in the United States and Canada alone making efforts to predict and control disease outbreaks as well as develop disease resistant soybean varieties of economic importance \citep{Allen2017}. %
However, today’s disease scouting and phenotyping techniques predominantly rely on human scouts and visual ratings. Human visual ratings are dependent on rater ability, rater reliability, and can be prone to human error, subjectivity, and inter/intra-rater variation \citep{Bock2010}. So there is a need for improved technologies for disease detection and identification beyond visual ratings in order to improve yield protection.
\par
Charcoal rot is an important fungal disease for producers in the United States and Canada, ranking among the top 7 most severe diseases in soybean from 2006 - 2014 and as high as the 2nd most yield limiting soybean disease in 2012 \citep{Allen2017,koenning2010suppression}. Charcoal rot has a large host range affecting other important economic crops such as corn, cotton, and sorghum making crop rotation a difficult management strategy \citep{short1980survival,Su2001}. %In addition there is limited chemical control measures leaving resistance breeding as an important approach to managing charcoal rot in soybean \citep{RomeroLuna2017}. In an effort to develop methods for earlier screening of charcoal rot resistance, one study proposed the cut-stem inoculation method for inoculating soybean seedlings in a growth chamber or greenhouse environment to measure lesion progression within a month after planting \citep{Twizeyimana2012}.% 
However, both field scouting for disease detection and small scale methods of charcoal rot evaluation still rely on visual ratings. These field and greenhouse screening methods for charcoal rot are time consuming and labor intensive. 
\par
Unlike visual ratings which only utilize visible(400-700 nm) region wavelengths, hyperspectral imaging can capture spectral and spatial information from wavelengths beyond human vision offering more usable information for disease detection. Hyperspectral imaging has been used for the detection and identification of plant diseases in barley, sugar beet, and wheat among others \citep{Bauriegel2011,Kuska2015,Mahlein2010}. In addition, hyperspectral imaging offers a potential solution to the scalability and repeatability issues faced with human visual ratings. Because of large data dimensions and redundancy, machine learning based methods are well suited to convert hyperspectral data into actionable information \citep{singh2016machine}. 
\par
In this work, we develop a supervised 3D-CNN based model  to learn the spectral and spatial information of hyperspectral images for classification of healthy and  charcoal rot infected samples. A saliency map based visualization method is used to identify the hyperspectral wavelengths that are most sensitive for the classification. We infer the importance of the wavelengths by analyzing the magnitude of saliency map gradient distribution of the  image across the hyperspectral wavelengths. To the best of our knowledge, this is the first work done on exploring the interpretation for the classification of hyperspectral data using saliency maps. This work is a societally relevant example of utilizing saliency maps to enable explanations of cues from hyperspectral data for disease identification. We are much more confident in the predictions of the model due to the physiologically meaningful explanations from the saliency  visualization. 
\section{Experiments}
 \subsection{Dataset}

%\begin{subfigure}[H]{0.3\textwidth}
%\centering

%\caption{}

%\end{subfigure}
%\begin{subfigure}[H]{0.3\textwidth}
%\centering
%\includegraphics[scale=0.3,width=3cm,height=2cm]{"./figures1/Picture9".png}
%\caption{}
%\label{fig 1b}
%\end{subfigure}
%\begin{subfigure}[H]{0.3\textwidth}
%\centering
%\includegraphics[scale=0.3,width=5cm,height=2cm]{"./figures1/Picture3".png}
%\caption{}
%\label{fig 1c}
%\end{subfigure}
   
Healthy and infected soybean stem  samples were collected at 3, 6, 9, 12, and 15 days after charcoal rot infection. Hyperspectral data cubes of the exterior of the infected and  healthy stems were captured at each time point of data collection prior to disease progression length measurements. The hyperspectral imaging system consisted of a Pika XC hyperspectral line imaging scanner, including the imager mounted on a stand, a translational stage, a laptop with SpectrononPro software for operating the imager and translational stage during image collection (Resonon, Bozeman, MT), and two 70-watt quartz-tungsten-halogen Illuminator lamps (ASD Inc., Boulder, CO) to provide stable illumination over a 400 – 1000 nm range. The Pika XC Imager collects 240 wavebands over a spectral range from 400 – 1000 nm with a 2.5 nm spectral resolution. The hyperspectral camera setup used in this study is shown in Figure 1a. Figure 1b shows an example of soybean stem captured at different hyperspectral wavelengths. The disease spread length from top of the stem is available for all the infected stem images which were used the for ground truth labeling of image classes. A reddish-brown lesion was developed on the stem due to charcoal rot infection often progressing farther down the inside of the stem than visible on the exterior of the stem. Figure 1c shows the RGB image of the disease progression comparison between interior and exterior region of a soybean stem.  The data-set contains 111 hyperspectral stem images of size 500x1600x240. Among the 111 images, 64 represent healthy stems and 47 represent infected stems. Data patches of resolution 64x64x240 pixels were extracted from the  stem images. The 64x64x240 image patches were applied as input to the 3D-CNN model. 
The training dataset consists of 1090 images. Out of 1090 training images, 940 images represent healthy stem and 150 images represent infected stem. All the images were normalized between 0 and 1. The validation and test dataset consists of 194 and 539 samples respectively. 
\begin{figure}[H]
\centering
\includegraphics[scale=0.5,width=12cm,height=4cm]{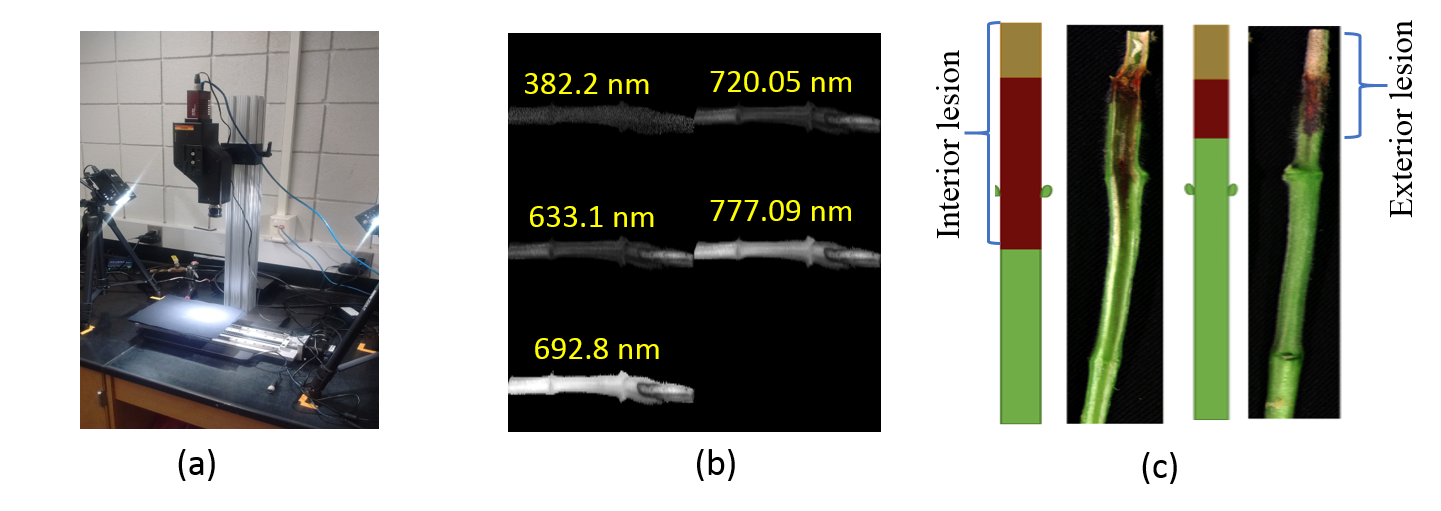}
\caption{ (a) Hyperspectral camera setup used in this study. (b) An example of a soybean stem imaged at different hyperspectral wavelengths. (c) RGB image of the disease progression comparison between interior and exterior region of soybean stem. Soybean stem was sliced in half , interior lesion length and  exterior lesion length were measured in mm.}
\label{fig:RGBIMAGE}
\end{figure} 

%\subsection{Hyperspectral Imaging}
%Hyperspectral data cubes of the exterior of the infected and  healthy stems were captured at each time point of data collection prior to disease progression length measurements. The Pika XC Imager collects 240 wavebands over a spectral range from 400 – 1000 nm with a 2.5 nm spectral resolution. The image was captured using the SpectrononPro software. Figure \ref{HYPER} shows the hyperspectral imaging setup used in the study and example of a soybean stem imaged at different wavelengths. 
%\begin{figure}
%\centering 
%\begin{subfigure}[t]{0.3\textwidth}
%\includegraphics[scale=0.50,width=4cm,height=4cm]{"./figures1/Picture9".png}
%\label{camset}
%\end{subfigure}
%\begin{subfigure}[t]{0.3\textwidth}
%\centering
%\includegraphics[scale=0.50,width=4cm,height=4cm]{"./figures1/Picture3".png}
%\label{hyperimage}
%\end{subfigure}
%\caption{Left: The hyperspectral imaging setup. Right: Examples of hyperspectral stem images at different wavelengths is shown in (b)}
%\label{HYPER}
%\end{figure}
\subsection{Model Architecture}
\begin{figure}[H]
\centering
\includegraphics[scale=0.8,width=12cm,height=5cm]{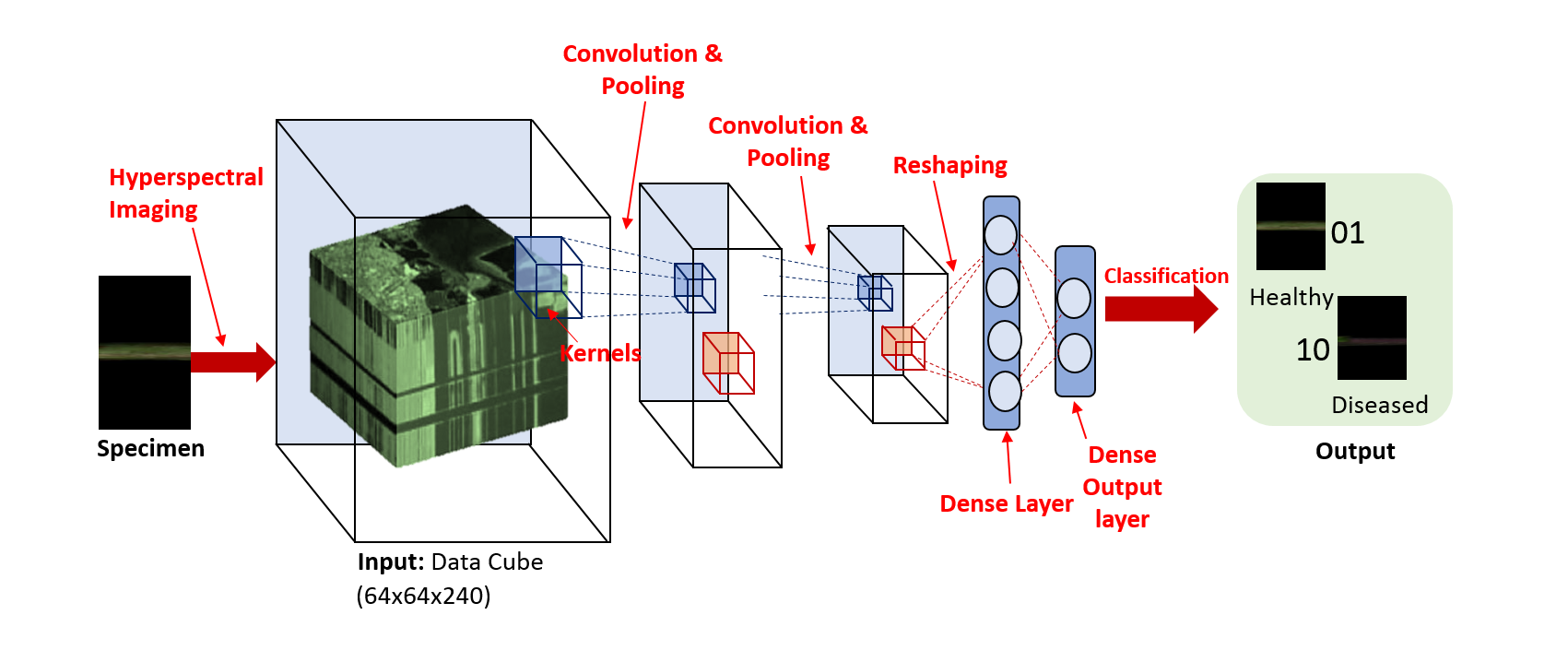}
\caption{3D Convolutional Neural Network Architecture for Classification.}
\label{fig:convnet}
\end{figure}

The 3D-CNN model consists of 2 convolutional layers interspersed with  2 max pooling layers followed by 2 fully connected layers. A relatively small architecture was used to prevent overfitting during training. Two  kernels of size 3x3x16 (3x3 in spatial dimension and 16 in spectral dimension) were used for convolving the input of the first convolution layer and four kernels of size 3x3x16 were used in the second convolution layer. Rectified Linear Input(RELU)  was used as the activation function for  the convolution output \citep{glorot2011deep}. A 2x2x2 maxpooling was applied  on the output of each convolutional layer.  Dropout with a probability of 0.25 was performed after first max pooling operation and with a probability of 0.5 after the first fully-connected layer. Dropout mechanism  was used to prevent overfitting during training \citep{Krizhevsky2012}. The first fully-connected layer consists of 16 nodes. The output of the second fully-connected layer (2 nodes) is fed to a softmax layer. 
Figure-\ref{fig:convnet} summarizes the 3D convolutional neural network architecture used in the study. 

\subsection{Training}
Adam optimizer was used to train our convolutional network weights on mini-batches of size 32 \citep{kingma2014adam}. We use a learning rate value $1$ x $10^-6$  and set $\beta_1$=0.9, $\beta_2$=0.999 and $\epsilon$ = $10^{-8}$. The convolution layer kernels were initialized with normal distribution with standard deviation of 0.05 and biases were initialized with zero. The dense layer neurons were initialized using glorot initialization \citep{pmlr-v9-glorot10a}. The 3D-CNN  model was trained for 126 epochs. Here, we use all the 240 wavelength bands of hyperspectral images for classification purpose. We trained using Keras \citep{chollet2015keras} with Tensorflow \citep{tensorflow2015-whitepaper} backend on a NVIDIA Tesla P40 GPU.
\subsection{Class Balanced Loss Function }
Because of imbalanced training data, weighted binary cross-entropy was used as a loss function. The loss ratio was 1:6.26 between more frequent healthy class samples and less frequent infected class samples. The class balanced loss improved our classification accuracy and F1-score.
\subsection{Classification Results}
We evaluate the learned 3D-CNN model on 539 test images. Table \ref{results} shows the classification results of the 3D CNN model. Classification accuracy of 95.73\% and precision value of 0.92 indicates a good generalizing capacity of the model for different infected stages of disease. The recall score is lower compared to precision score  as  disease severity indications are very sparse in some locations of the exterior stem region. The F1-score of infected class of the test data was 0.87.
%\begin{figure}
%\centering 
%\begin{subfigure}[b]{0.3\textwidth}
%\includegraphics[scale=0.5,width=5cm,height=4cm]{"./figures1/Picture5".png}
%\caption{}
%\label{fig:acc}
%\end{subfigure}
%\begin{subfigure}[b]{0.3\textwidth}
%\centering
%\includegraphics[scale=0.5,width=5cm,height=4cm]{"./figures1/Picture6".png}
%\caption{}
%\label{fig:loss}
%\end{subfigure}
%\caption{Left: Training and validation accuracy comparisons. Right: Training and validation loss comparisons}
%\label{fig:accloss}
%\end{figure}  

%\begin{table}[t]
%  \caption{Output Confusion Matrix}
%  \label{cm}
%  \centering
%  \begin{tabular}{llll}
%    \toprule
%    True Positive & True Negative & False Positive & False Negative     \\
%    \midrule
%    74 & 441 & 3 & 21   \\
%    \bottomrule
%  \end{tabular}
%\end{table}

\begin{table}[H]
  \caption{Classification Results}
  \label{results}
  \centering
  \begin{tabular}{llll}
    \toprule
    Precision & Recall & F1-Score & Classification Accuracy     \\
    \midrule
    0.92 & 0.82 & 0.87 & 95.73   \\
    \bottomrule
  \end{tabular}
\end{table}
\subsection{Saliency Map Visualization}
\begin{figure}[H]
\centering
\includegraphics[scale=0.5,width=12cm,height=6cm]{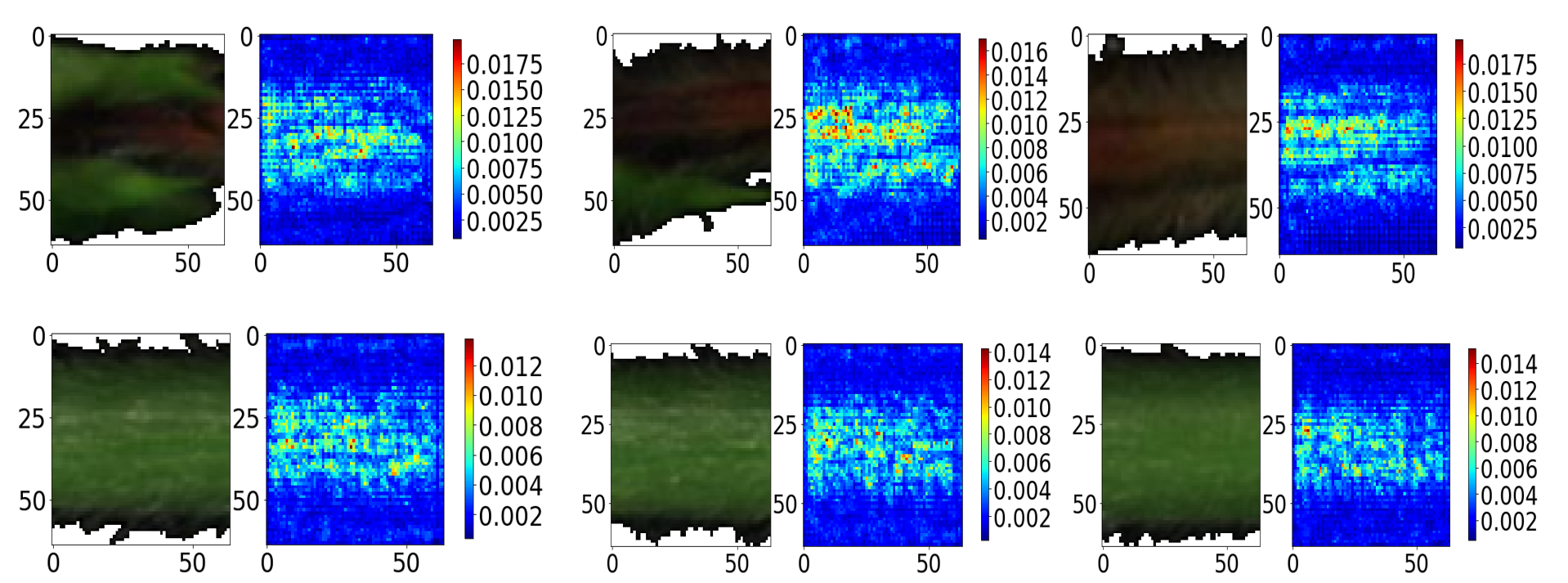}
\caption{Image specific class saliency maps for the infected(Top) and healthy(Bottom) test images. The magnitude of the gradient of the maximum predicted class score with respect to the input image in the visualizations illustrate the sensitivity of the pixels to classification.}
\label{smap}
\end{figure} 

We visualize the parts of the image that were most sensitive to the classification using image specific class saliency map. Specifically, the  magnitude of the gradient of the maximum predicted class score with respect to the input image was used to  identify the most sensitive pixel locations for classification \citep{simonyan2013deep}. The saliency map visualizations of the healthy and infected samples are shown in Figure \ref{smap}. The magnitude of gradient of each pixel indicates the relative importance of the pixel in the prediction of the output class score. The saliency maps of the infected stem images has high magnitude of gradient values in locations corresponding to the severely infected regions(reddish-brown) of the image as shown in the Figure \ref{smap}. This indicates that the severely infected regions of the image contains the most sensitive pixel locations for prediction of the infected class score. For both the healthy and infected images, the saliency map gradients  were concentrated around the mid region of the stem.

\subsection{Explaining importance of hyperspectral wavelengths for classification using saliency maps}

Let $I_1,I_2,...I_N$ be the $N$ test images for disease classification. Let $W$ be the gradient of the maximum predicted class score with respect to the input image. Each pixel $(x,y)$ in the image $I_i$ is maximally activated by one of the  240 wavelength channels. Let us assume that the element index of $W$ corresponding to a pixel location $(x,y)$ in wavelength channel $C$ of a image $I_i$ is $g(x,y,C)$. For each pixel location $(x,y)$ in image $I_i$, let $C^\textbf{*}$ be the wavelength with maximum magnitude of $W$ across all channels.    
\begin{equation}
C^\textbf{*}=\argmax_{C{\in}{(1,2,...240)}}|W_{g(x,y,C)}|\; \textbf{For}\;(x,y) \in I_i 
\end{equation}

Note, that $C^\textbf{*}$ is a function of $(x,y)$. The histogram of $C^\textbf{*}$ from all pixel locations of the $N$ test images is shown in Figure \ref{activations}. It illustrates the percentage of pixel locations from all $N$ test images with maximum saliency gradient magnitude from each wavelength. The importance of a specific hyperspectral wavelength can be quantified by the percentage of pixels with  maximum saliency gradient magnitude in that wavelength. The histogram reveals several important aspects of our model. First, wavelength 733 nm ($C^\textbf{*}$=130)  from the near-infrared region was the most sensitive among all wavelengths in the test data. Second, the 15 wavelengths in the spectral region of 703 to 744 nm were responsible for maximum magnitude of gradient values in 33\% of the pixel locations of the test image. Further, the wavelengths in the visible region of the spectrum (400-700 nm) were more sensitive for the infected samples compared to the healthy samples.  
\begin{figure}[H]
\centering
\includegraphics[scale=0.3,width=9cm,height=7cm]{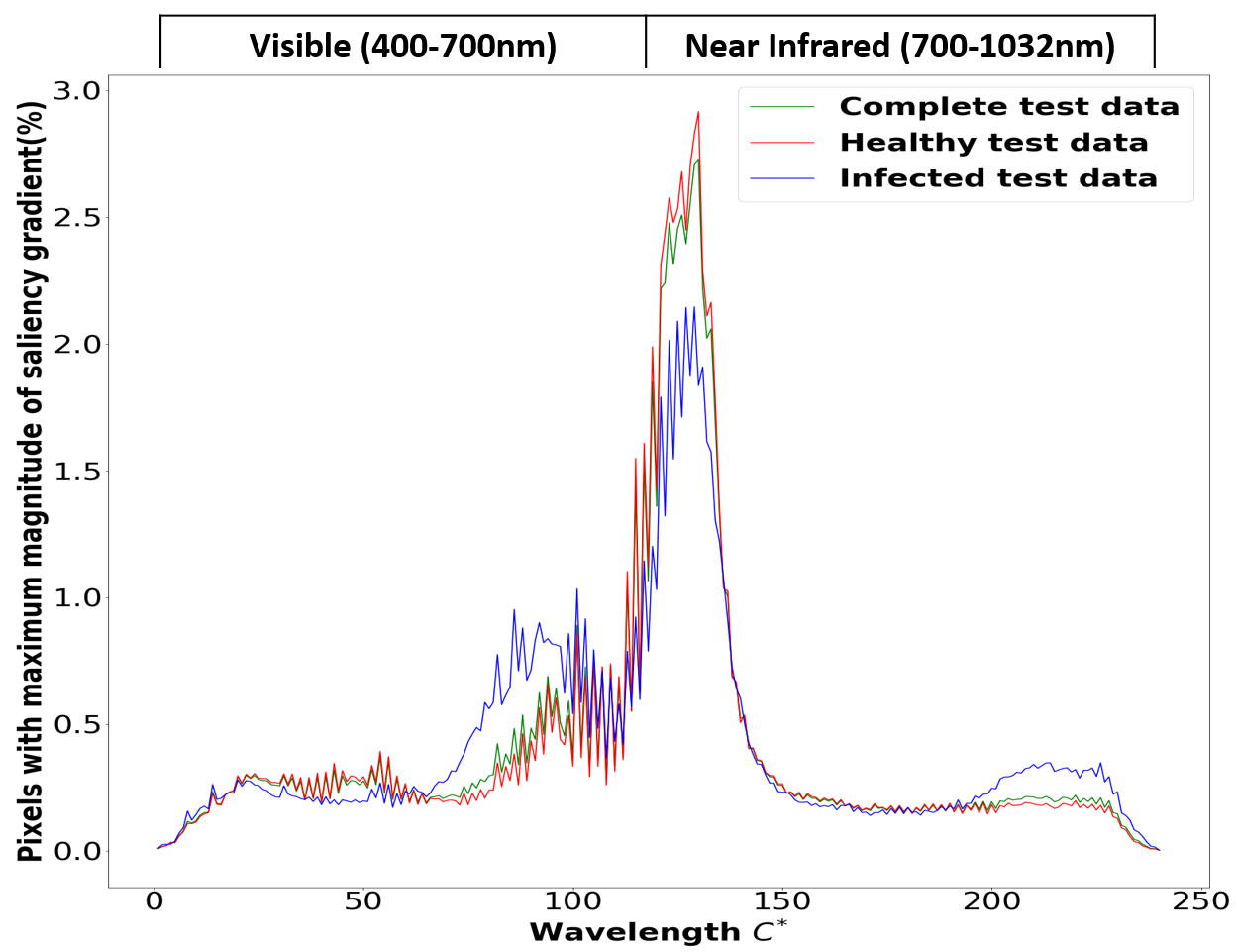}
\caption{Histogram of $C^\textbf{*}$ from all the test images. It illustrates the percentage of pixel locations from all $N$ test images with maximum magnitude of saliency gradient from each wavelength.}
\label{activations}
\end{figure}
The magnitude of the gradient of maximum predicted class score with respect to the the most sensitive wavelength channel $C^\textbf{*}=130$ (733 nm) of the test data is shown in Figure \ref{fig:130}. Wavelength  specific data gradient magnitude reveals the sensitivity of pixels in that wavelength for classification. Hence, wavelength specific visualization is helpful in understanding the importance of a individual wavelengths for classification.
\begin{figure}[H]
\centering
\includegraphics[scale=0.5,width=8cm,height=3cm]{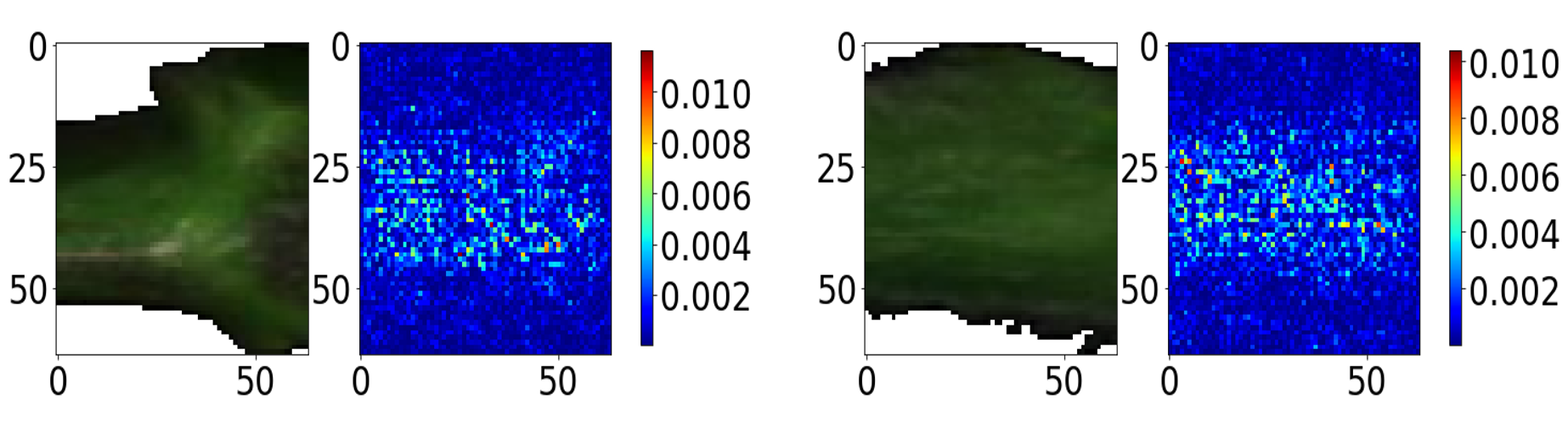}

\caption{Visualizations illustrate the  magnitude of the gradient of the maximum predicted class score with respect to the most sensitive wavelength $C^\textbf{*}=130$ (733 nm) of the complete test data. $C^\textbf{*}=130$ is the wavelength  which has maximum saliency  gradient magnitude at most number of pixel locations in the complete test data.}
\label{fig:130}
\end{figure}
\section{Conclusions}
We have demonstrated that a 3D CNN model can effectively be used to learn the high dimensional hyperspectral data to identify charcoal rot disease in soybeans.  We show that a saliency map  visualization can be  used to explain the importance of hyperspectral wavelengths in classification. Thus, using saliency map enabled interpretability, the model was able to track the physiological insights of the predictions. Hence, we are more confident of the predictive capability of our model. In future, band selection based on robust interpretability mechanisms  will  be helpful in dimensionality reduction of the large hyperspectral data and also in designing a multispectral camera system for high throughput phenotyping.

%\section*{References}
\bibliographystyle{plainnat}
\bibliography{E:/3DCNNcharcoalrot/3DCNNpaperskeleton/nipsworkshop}

\end{document}